\documentclass[letterpaper, 10 pt, conference]{ieeeconf}

\IEEEoverridecommandlockouts

\usepackage{graphics} \usepackage{times} \usepackage{amsmath} \usepackage{amssymb}  \usepackage{amsfonts}
\usepackage{subfigure}
\usepackage{hyperref}
\usepackage{graphicx}
\usepackage{float}
\usepackage{caption}
\usepackage{multirow}
\usepackage{color}
\usepackage{xcolor}
\usepackage{booktabs}

\newcommand{\ie}{\textit{i}.\textit{e}., }

\title{\LARGE \bf
FSGlove: An Inertial-Based Hand Tracking System with Shape-Aware Calibration
}

\author{Yutong Li$^{1,*}$, Jieyi Zhang$^{1,*}$, Wenqiang Xu$^{1}$, Tutian Tang$^{1}$ and Cewu Lu$^{1,\dagger}$\thanks{$^*$ indicates equal contribution.}
\thanks{$\dagger$ Corresponding Author. }
\thanks{$^{1}${\tt\small \{yutong, yi\_eagle, vinjohn, tttang, lucewu\}@sjtu.edu.cn}. Yutong Li, Jieyi Zhang, Wenqiang Xu, and Tutian Tang are with School of Electronic Information and Electrical Engineering, Shanghai Jiao Tong University, Shanghai, China, and also with the Meta Robotics Institute, Shanghai Jiao Tong University, Shanghai, China. Cewu Lu is the corresponding author, a member of Qing Yuan Research Institute and MoE Key Lab of Artificial Intelligence, AI Institute, Shanghai Jiao Tong University, China. }
\thanks{This work was supported by the Shanghai Commitee of Science and Technology, China (No. 24511103200), by the National Key Research and Development Project of China (No. 2022ZD0160102), by the Fundamental Research Funds for the Central Universities (project number YG2022ZD003), Shanghai Artificial Intelligence Laboratory, XPLORER PRIZE grants.}}
\begin{document}

\maketitle
\thispagestyle{empty}
\pagestyle{empty}
\renewcommand{\thefootnote}{}

\begin{abstract}
Accurate hand motion capture (MoCap) is vital for applications in robotics, virtual reality, and biomechanics, yet existing systems face limitations in capturing high-degree-of-freedom (DoF) joint kinematics and personalized hand shape. Commercial gloves offer up to 21 DoFs, which are insufficient for complex manipulations while neglecting shape variations that are critical for contact-rich tasks. We present FSGlove, an inertial-based system that simultaneously tracks up to 48 DoFs and reconstructs personalized hand shapes via DiffHCal, a novel calibration method. Each finger joint and the dorsum are equipped with IMUs, enabling high-resolution motion sensing. DiffHCal integrates with the parametric MANO model through differentiable optimization, resolving joint kinematics, shape parameters, and sensor misalignment during a single streamlined calibration. The system achieves state-of-the-art accuracy, with joint angle errors of less than 2.7$^\circ$, and outperforms commercial alternatives in shape reconstruction and contact fidelity. FSGlove’s open-source hardware and software design ensures compatibility with current VR and robotics ecosystems, while its ability to capture subtle motions (e.g., fingertip rubbing) bridges the gap between human dexterity and robotic imitation. Evaluated against Nokov optical MoCap, FSGlove advances hand tracking by unifying the kinematic and contact fidelity. Hardware design, software, and more results are available at: \url{https://sites.google.com/view/fsglove}.

\end{abstract}

\section{Introduction}
The human hand, a remarkably dexterous end-effector capable of executing intricate tasks, serves as both the inspiration and benchmark for robotic hand design and manipulation research. Accurate motion capture of the hand during manipulation is critical for collecting data essential to diverse downstream applications, including hand pose estimation, teleoperation, and imitation learning. Achieving high-fidelity hand motion capture hinges on two factors: joint kinematics and hand shape modeling.

\begin{figure}
    \centering
    \includegraphics[width=0.8\columnwidth]{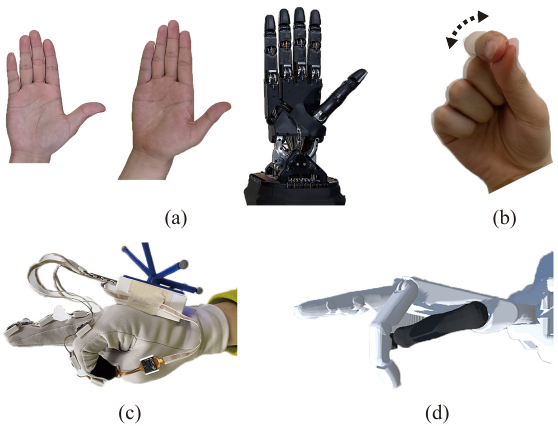}
    \caption{\textbf{Hand shape variation.} (a) The variation in hand shapes across different human subjects and a robotic hand. (b) Thumb-index fingertip rubbing requires a slight twist DoF.  (c), (d) Motion transfer often fails when adapted between hands with heterogeneous shapes, even when all the joints are mapped.}
    \label{fig:intro}
\end{figure}

Currently, the most comprehensive commercially available hand MoCap glove \cite{cyberglove} can capture at most 21 degrees of freedom (DoFs). However, complex manipulations, such as thumb-index fingertip rubbing, are still beyond the ability to capture, as they require additional torsional DoFs in the proximal or middle phalanges that are often overlooked (Fig. \ref{fig:intro}), but recently, some dexterous robotic hands \cite{leaphand} tend to support control such DoFs. Similarly, hand shape variations, driven by differences in bone length and soft-tissue composition, demand personalized models for accurate motion reconstruction, particularly during contact-rich tasks involving objects or self-interaction. Existing hand motion capture systems, whether commercial \cite{manus, vrtrix} or research-oriented \cite{glove1, glove2, glove3}, focus primarily on joint angle estimation with limited DoFs and neglect the shape variations. This oversight hinders the transfer of in-manipulation motions to virtual avatars or anthropomorphic robotic hands, where both kinematic and contact fidelity are paramount (Fig. \ref{fig:intro}).

To bridge this gap, we present FSGlove, a novel inertial-based hand tracking system simultaneously capturing up to 48-DoF joint kinematics and personalized hand shape. Each finger joint and the dorsal region are instrumented with inertial measurement units (IMUs), enabling high-resolution motion sensing. A key innovation is DiffHCal, a shape-aware calibration method that embeds personalized hand-shape estimation into a streamlined calibration protocol. By leveraging the parametric MANO hand model \cite{mano}, DiffHCal aligns captured sensor data to MANO’s joint and shape parameters through a differentiable optimization process. This framework uses a series of predefined reference poses to simultaneously resolve skeletal joint parameters, shape parameters, and manual installation errors in a single optimization process. Moreover, this process integrates seamlessly into standard glove calibration workflows, requiring no additional steps compared with commercial systems while delivering contact-consistent hand models. To democratize access, we open-source the hardware design, low-level drivers for sensor calibration, and high-level interfaces for integration with motion capture ecosystems (e.g., OptiTrack, HTC Vive), enabling plug-and-play compatibility with robotics and VR frameworks.

We evaluate FSGlove across four metrics: (1) joint angle accuracy ($\leq 2.7^{\circ}$), (2) shape reconstruction precision (mean mesh error $\leq 3.6$mm), (3) fingertip tracking error (mean positional error $\leq 16$mm), and (4) hand-object interaction fidelity (mean positional error $\leq 20$mm), benchmarking against commercial glove systems. Despite its low-cost design, FSGlove achieves state-of-the-art performance in shape-aware tracking, surpassing commercial alternatives (e.g., Manus Metaglove Pro, VRTRIX) in contact-rich manipulation tasks.

We summarize our contributions as follows:
\begin{itemize}
    \item FSGlove, the first open-source, high-DoF (up to 48 DoFs) data glove integrating inertial sensing with shape-aware calibration.
    \item DiffHCal, a differentiable calibration framework that infers personalized hand shape during standard wear-time procedures.
    \item Comprehensive validation demonstrating high-precision accuracy in joint measurement and shape reconstruction.
\end{itemize}

\section{RELATED WORKS}
\subsection{Data Glove Systems}
The development of data gloves dates back to the 1970s. Since then, various types based on different sensors have emerged, each with distinct pros and cons.
The early Sayre Glove in 1977 \cite{sayreglove} used flexible tubes and photocells. It could measure finger bending but had few sensors, was hard-wired and cumbersome, and had limited applications. In the 1980s, the Data Glove \cite{dataglove} was a significant advancement. It used plastic tubes and light-based sensors for joint angle measurement, followed by the Power Glove~\cite{powerglove} and the Super Glove~\cite{glove_survey}. These gloves could measure finger joint bending but the cloth support restricted hand movement, affecting measurement accuracy. They also needed complex calibration, and some had problems such as non-linear mapping and sampling limitations \cite{glove_survey}.

MemGlove\cite{resistiveglove} combined resistive and fluidic sensors for hand pose reconstruction. However, it featured a complex fabrication and sensor integration process.
Gloves with flex sensors~\cite{flexglove} were inexpensive and easy to use for measuring finger flexion. However, the sensors were sensitive to temperature and vulnerable to wear and tear. In addition, the large size of the sensors limited the degrees of freedom they could capture.

With advancements in microelectronics, gloves equipped with MEMS-based inertial measurement unit (IMU) sensors have emerged. The standardized production of sensor chips has significantly reduced costs, enabling mass production of these gloves. These gloves \cite{imuglove1, imuglove2} leverage IMU characteristics to provide superior dynamic responses, enabling comprehensive capture of hand and arm movements. \cite{glove2} introduced a data glove prototype incorporating flexible printed circuits (FPC) and proposed the use of angle sensors to verify IMU accuracy. Similarly, \cite{lin_design_2018} also presented a modular IMU-based data glove design. However, these approaches neglected adaptability to varying hand sizes and lacked optimization for machine learning applications. Nonetheless, these sensors face calibration challenges \cite{imuglove2} and occasional susceptibility to external magnetic field interference.

\subsection{Glove Calibration Methods}
To alleviate or address problems caused by glove sensors, data glove calibration research has seen continuous development in recent years.
Chou et al. \cite{calibration1} used linear regression for Cyberglove calibration, relying on a vision system to handle the complex relationship between sensor readings and joint angles. Kahlesz et al. \cite{calibration2} modeled sensor cross-couplings for high-degree DoF data gloves like the Cyberglove, achieving visually appealing calibrations without extra hardware design. Sun et al. \cite{calibration3} applied genetic algorithms to calibrate exoskeleton data gloves, accounting for hand-size and wearing-position variations.
Zhou et al. \cite{calibration4} simplified CyberGlove calibration in virtual rehabilitation via artificial neural networks, benefiting disabled patients. Connolly et al. \cite{calibration5} improved the accuracy and usability of the 5DT Data Glove 14 Ultra using neural networks, eliminating the need for traditional calibration when the joint mobility is limited.

\section{System Architecture and Design}

\begin{figure}
    \centering
    \includegraphics[width=0.9\columnwidth]{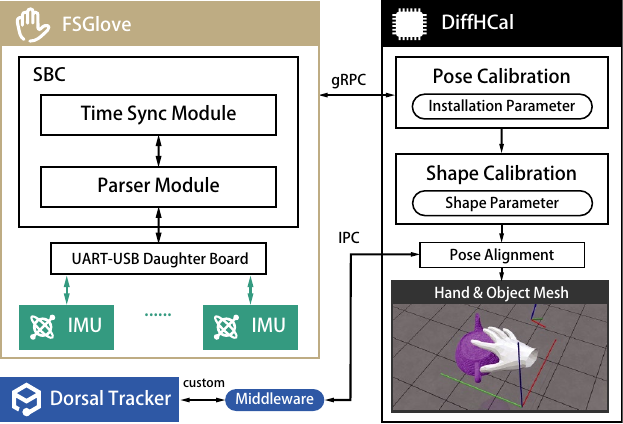}
    \caption{\textbf{System Architecture.} The system gathers IMU sensor data via an SBC and then sends them to the DiffHCal framework. The DiffHCal framework calibrates and models the hand shape, aligns the data, and outputs a final hand-and-object mesh.}
    \label{fig:glove_diagram}
\end{figure}

For both robotics and motion capture applications, stable operation and real-time processing of data from multiple sensors are essential. To meet this requirement, this study includes the distributed motion capture system shown in Fig. \ref{fig:glove_diagram}. The system comprises three hardware-independent subsystems: (1) the FSGlove for finger kinematics recording, (2) a dorsal tracker for global hand translation, and (3) a data acquisition suite for sensor fusion and visualization. These components communicate via Ethernet or Wi-Fi with synchronized clocks. We detail the FSGlove hardware in Sec. \ref{sec:glove_hardware}, the dorsal tracker in Sec. \ref{sec:dorsal_tracker}, and the data acquisition pipeline in Sec. \ref{sec:data_suite}.

\begin{figure}
    \centering
    \includegraphics[width=0.85\columnwidth]{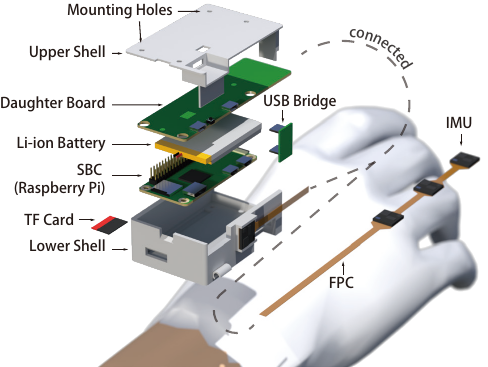}
    \caption{\textbf{Glove hardware design.} The Raspberry Pi Zero 2W serves as the system’s core, receiving sensor data from a custom UART-USB daughter board over a USB bridge. The daughter board collects IMU data via a flex-printed circuit, and the IMUs are attached to the glove surface.}
    \label{fig:glove_layout}
\end{figure}

\begin{figure}
    \centering
    \includegraphics[width=0.9\columnwidth]{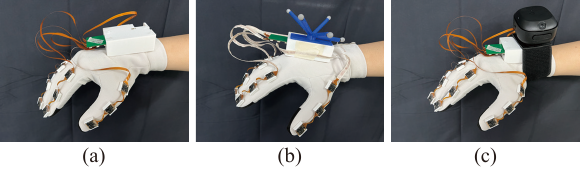}
    \caption{\textbf{FSGlove's board compatibility.} (a) Basic FSGlove without a dorsal tracker, capturing only finger kinematics.
    (b) FSGlove is equipped with an optical dorsal tracker and tapped FPC, which is integrated with the Nokov MoCap system. (c) FSGlove is equipped with an HTC Vive Tracker, integrated with the Vive VR system.}
    \label{fig:glove_photo}
\end{figure}

\subsection{Hardware Setup of FSGlove}\label{sec:glove_hardware}
\paragraph{Topology and Mechanical Design} 
Human finger joints function as hinge-like constraints, permitting rotations between hierarchically connected phalanxes (proximal $\to$ middle $\to$ distal). To capture this kinematic chain, we deploy three IMUs per finger, positioned dorsally on the proximal, middle, and distal phalanxes, supplemented by a dorsum-mounted IMU for global orientation. This forms a 16-IMU network (5 fingers × 3 + 1 dorsum).

To minimize wiring complexity while preserving dexterity, IMUs are soldered onto a custom FPC.
The FPC is bonded to a stretchable nylon glove using 3D-printed mounts to prevent sensor slippage.
A dorsal enclosure housing the single-board computer (SBC) and a custom daughter board moves synchronously with the hand.

\paragraph{IMU Selection}  We selected the HI229 module \cite{hi229} (BNO055 chip) for its compact size, cost, and accuracy. It delivers fused orientation estimates with 0.8$^\circ$ static angular error and 2.5$^\circ$ dynamic error at up to 400 Hz sample rates – performance comparable with commercial gloves\cite{vrtrix}.

\paragraph{Single-Board Computer (SBC)} A Raspberry Pi 2W (Pi2W) is used for its quad-core $1.0$GHz CPU and integrated Wi-Fi, which are sufficient for processing data from 16 IMUs and transmitting it wirelessly. As an off-the-shelf component, the Pi2W ensures system replicability.

\paragraph{USB-UART Daughter Board} 

To overcome the Pi2W's port limitations, we designed a daughter board with dual CH9344 8-Port USB-UART bridge chips. This board expands connectivity to 16 UART ports, maintaining a $480$ Mbps uplink to the Pi2W through a custom USB bridge, all housed within a 3D-printed case to prevent electrical shorts.

\paragraph{Battery System for Mobility} A $1100$ mAh Li-ion battery with charge-management circuitry powers the Pi2W, enabling untethered operation. With default configuration, the glove can operate continuously for 2 hours. This balance between runtime and glove weight (110g) ensures usability during prolonged manipulation tasks.

\paragraph{Cost Analysis} The total bill-of-materials cost for the 16-IMU glove configuration is $\$426$, including assembly expenses (detailed breakdown in Supplementary). IMUs account for 75\% of this cost (\$320), indicating that significant cost reductions are possible for other applications requiring fewer sensors.

\subsection{Dorsal Tracker}\label{sec:dorsal_tracker}
While IMUs capture hand joint orientations, global translational motion requires complementary tracking. We adopt the Nokov optical system to resolve 6DoF dorsal poses.As shown in Fig. \ref{fig:glove_photo}, a 3D-printed, tree-like structure with infrared (IR) reflective markers is mounted on the glove for stable tracking. The FPC is shielded with tape to reduce IR reflection noise.

\subsection{Data Acquisition Suite}\label{sec:data_suite}
The data acquisition suite integrates synchronized streams from the FSGlove and the dorsal tracker. We detail four critical design aspects below.

\paragraph{Communication Protocols} 
We use gRPC for low-latency, bidirectional data transmission, and its integrated Protobuf protocol simplifies serialization. For dorsal tracking, a middleware layer uses inter-process communication (IPC) to integrate with various tracking systems, minimizing latency and computational overhead.

\paragraph{Time Synchronization} 
To synchronize the entire system, SBC periodically aligns its clock with a PC time-server via the Network Time Protocol (NTP). While the Precision Time Protocol (PTP) is recommended for synchronizing with third-party dorsal tracking systems, the Nokov system we use employs custom protocols for synchronization. The 16 onboard IMUs are synchronized based on timestamped data packet arrival using an approximate policy inspired by the Robot Operating System (ROS) message filter.

\paragraph{The Data Flow} 

The SBC and PC communicate over Wi-Fi. The SBC processes raw byte streams from 16 IMUs sampled at $100$ Hz, converting them into single entries containing triaxial acceleration, angular velocity, and fused orientation. This sample rate is a result of the trade-off between battery consumption, computational load, and motion fidelity. These entries are streamed to the PC, where a calibration module integrates dorsal tracker poses to compute the final hand pose, as detailed in Sec. \ref{sec:pose_calibration}.

\paragraph{Delay Analysis} 
System latency measurements quantify a total glove delay of $24$ ms, predominantly caused by Wi-Fi channel congestion at the SBC. whereas the wired Nokov tracker has a latency of just 1 ms.Middleware and memory operations introduce negligible overhead ($<0.5$ ms). GPU-accelerated (NVIDIA 2080 Ti) reconstruction enables real-time visualization at 25 Hz, resulting in a final display latency of 40 ms.

\section{MoCap Data Alignment and Shape-aware Calibration}
In the FSGlove-based MoCap system, it is necessary to convert the raw sensor readings to hand motion data. We propose a unified approach, DiffHCal, to align and calibrate the data from the MoCap system and consider both joints and shapes.
DiffHCal is a end-to-end processing pipeline built on top of the MANO hand model \cite{mano}, which is widely utilized in many hand-related tasks and has a rich ecosystem. Importantly, it parameterized both the joint and shape of the hand. The architecture of DiffHCal is illustrated in Figure \ref{fig:glove_diagram}. It consists of pose calibration module, shape calibration module and support alignment with dorsal tracking systems.

\subsection{Pose Alignment and Calibration}\label{sec:pose_calibration}
The raw IMU data captured by the glove does not represent joint angles but the phalange rotations. Therefore, it is necessary to first convert them to joint angles.
Then, considering the glove's wearing, the IMU may have different relative poses to the corresponding finger phalange for each glove's wearing process. Such installation poses should be calibrated each time the glove is worn. 

\paragraph{Pose Alignment}
We first establish a virtual world coordinate system $\mathcal{M}$ for MANO. IMU has a uniform West-North-Up (WNU) coordinate system as the world coordinate system $\mathcal{W}$.
IMUs are indexed according to the finger link ID on MANO. Ideally, we assume that each IMU is installed and aligned with the link frame perfectly. $R_i^{\mathcal{W}}$ is $i$-th IMU's reading while $R_i^{\mathcal{M}}$ represents the rotation of $i$-th finger link in the $M$ coordinate system. Then, rotation from the WNU world coordinate system to the MANO's world coordinate system is $A$:
\begin{equation}\label{eq:transform}
    R^\mathcal{M}_i = AR^\mathcal{W}_i.
\end{equation}

By defining the joint index with its parent link's index in MANO's convention, for each finger, joint $i$ connects link $i$ and link $i-1$. Therefore, the joint angle is defined as:
\begin{equation}\label{eq:joint}
      \theta_i = R^\mathcal{M}_i  (R^\mathcal{M}_{i-1})^{-1}. 
\end{equation}

\paragraph{Pose Calibration} 
In Eq. \ref{eq:transform} and Eq. \ref{eq:joint}, $R^\mathcal{W}_i$ is read from IMU sensor, $R^\mathcal{M}_i$ is read from the virtual environment where the MANO exists (\ie a 3D simulator, such as RFUniverse \cite{rfu}). Only the transformation matrix $A$ is needed.

The challenge in calculating $A$ is how a human in the real world can pose the hand the same way as the virtual hand. Since $A$ is the same for all the finger links, one finger link pose pair is all we need.

However, in practice, the IMU cannot be perfectly aligned with the link frame, and its orientation changes each time the glove is worn because it is flexible.
Thus, considering such installation error, Eq. \ref{eq:transform} should be:
\begin{equation}
    R^\mathcal{M}_i C_i = AR^\mathcal{W}_i,
\end{equation}
$C_i$ is the correction rotation indicating the installation error from $i$-th link frame.

To solve both $A$ and $C_i$, at least two sets of standard reference hand pose should be defined, and the least square method can be applied.

\begin{figure}
    \centering
    \includegraphics[width=0.9\columnwidth]{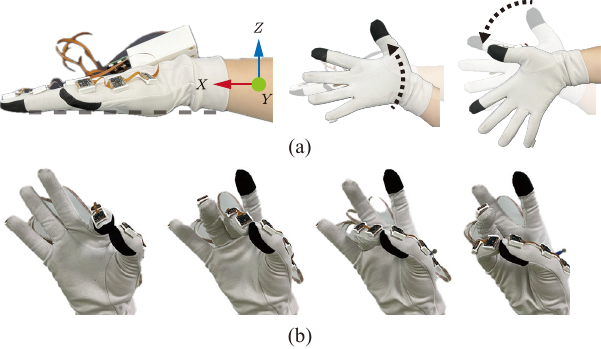}
    \caption{\textbf{Calibration Poses.} (a) The pose calibration process involves only three simple poses. (b) The shape calibration process requires each fingertip to touch the thumb tip.}
    \label{fig:calibration_desc}
\end{figure}

In practice, we define three standard reference poses: the rest pose, the $x$-axis rotation pose, and the $y$-axis rotation pose. As shown in Fig. \ref{fig:calibration_desc}.(a),
The rest pose is when flatting the hand palm and face downward onto a surface, and spreading all fingers as wide as possible.
The $x$-axis rotation pose is when we rotate the wrist of the rest pose around the $x$-axis (keeping the fingers still and orienting the palm vertically).
The $y$-axis rotation pose is when we rotate the wrist of the $x$-axis rotation pose around the $y$-axis (keeping the fingers still while moving the palm in a vertical plane).

\subsection{Shape Calibration} \label{sec:shape_calibration}
When two people try to manipulate an object with the same hand pose, the one with the bigger hand is already in contact with the object, while the one with the smaller hand is not. Failing to consider the shape variation will cause severe problems when downstream tasks (e.g., VR/AR, manipulation data collection) involve object interaction.
Therefore, we also need to calibrate the shape size.

To personalize different human hand shapes, we use the MANO shape parameter vector $\beta$. Given pose parameters $\theta$ and shape parameters $\beta$, the hand mesh model is defined as
\begin{equation}
    (V, E) = M(\theta, \beta) ,
    \label{eq:mano}
\end{equation}
where $V$ denotes the vertex set, $E$ denotes the edge set. 

One way to personalize the hand shape is to scan the hand with a glove and apply the hand rig to the scanned model. It would require considerable modeling time and is not practical from the user end. Another way is to use a personalized hand reconstruction method such as \cite{harp}, but such a method is not trained on the hand with gloves.

Considering that shape variation mostly influences the contact states, we propose calibrating the hand shape using a few touch-based standard reference poses. 
In the standard pose, we will define a contact state $c_{jk}=(v_j, v_k)$, the contact set $\mathcal{C}=\{c_{jk}\}$, $v_j$ is the $j$-th vertex. As shown in Fig. \ref{fig:calibration_desc}. (b), in a thumb-index pinch pose, if the thumb is in contact with the index finger, $744$-th vertex and $320$-th vertex should be in contact, where indices are of MANO's convention.
Given the predefined contact relationship in the reference poses, and humans try to replicate the reference pose in the real world, we apply the measured hand pose to the MANO hand and define the shape error metric as:
\begin{equation}
    E_{shape}(\beta) = \sum_{c_{jk}\in\mathcal{C}} \Vert v_{j}(\theta, \beta) - v_k(\theta, \beta)\Vert_2^2.
\end{equation}
Since $M$ in Eq. \ref{eq:mano} is differentiable \cite{mano}, minimizing $E_{shape}$ to find the optimal $\beta$ is also differentiable.

Considering most manipulation applications are grasping-based, optimizing the contact states on the fingertips would suffice for such scenarios. It is the default setting for our experiments.
If more complex manipulations are considered, such as in-hand manipulation, the contact states on another part of the hand could also be considered, which is trivial to extend with the same optimization scheme.

\subsection{Alignment with the Dorsal Tracking System}
Finally, the wrist pose of the personalized MANO must be obtained. We mentioned before that in the FSGlove-based MoCap system, the dorsal rotation is obtained from IMU, and the dorsal translation is obtained from the external dorsal tracker.
The alignment and calibration process is similar to the pose part. The main difference is that we should calculate the transformation matrix $A'$ between the dorsal tracker's world coordinate system and MANO's.

Note that we can use the same reference hand poses as the ones described in Sec. \ref{sec:pose_calibration} to accomplish the alignment process. Thus, we can simultaneously accomplish the pose alignment process for the hand and wrist if necessary. In this way, all the calculations for hand pose, shape, and wrist pose are differentiable.

\section{Evaluation Experiments}

To evaluate the performance of these gloves, we designed a series of experiments based on common research and application scenarios and selected multiple different hand pose estimation technologies for comparative study. They include the VRTRIX MoCap glove, which also uses an IMU solution, the Meta Quest3, which adopts a vision-based hand recognition algorithm, and the Manus Metaglove Pro based on magnetic sensing, which claims to be the most precise hand tracking solution. Details of their information can be found in Tab. \ref{tab:comparison}.

\begin{table}[!t]
\caption{\textbf{Comparison of hand glove motion capture solutions.} Our glove offers the lowest cost while maintaining the sampling rate and pose DoFs.}
\label{tab:comparison}
\centering
\begin{tabular}{l | c c r}
\toprule
\textbf{Solution} & \textbf{Technology} & \textbf{Sampling Rate (Hz)} & \textbf{Price} \\
\midrule
Ours (FSGlove) & IMU (16 unit) & 100 & \textbf{\$426} \\
VRTRIX Pro & IMU (6 unit) & \textbf{180} & \$1,400 \\
Meta Quest 3 & Vision-Based & 60 & \$499 \\
Manus Metaglove & Magnetic Sensor & 120 & \$5,250 \\
\bottomrule
\end{tabular}
\end{table}

\subsection{Single Joint Measurement}
\label{sec:single-joint-measurement}

\begin{figure}
    \centering
    \includegraphics[width=0.9\columnwidth]{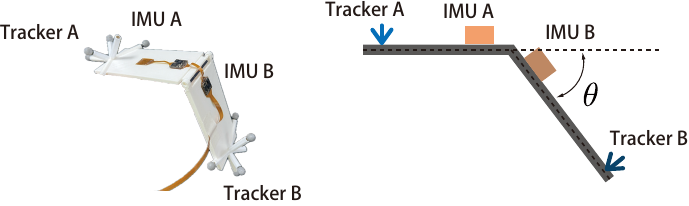}
    \caption{\textbf{Settings of single joint error experiments.} We use optical system as ground truth to test the single joint error.}
    \label{fig:experiment_0_setting}
\end{figure}

\begin{figure}
    \centering
    \includegraphics[width=0.9\columnwidth]{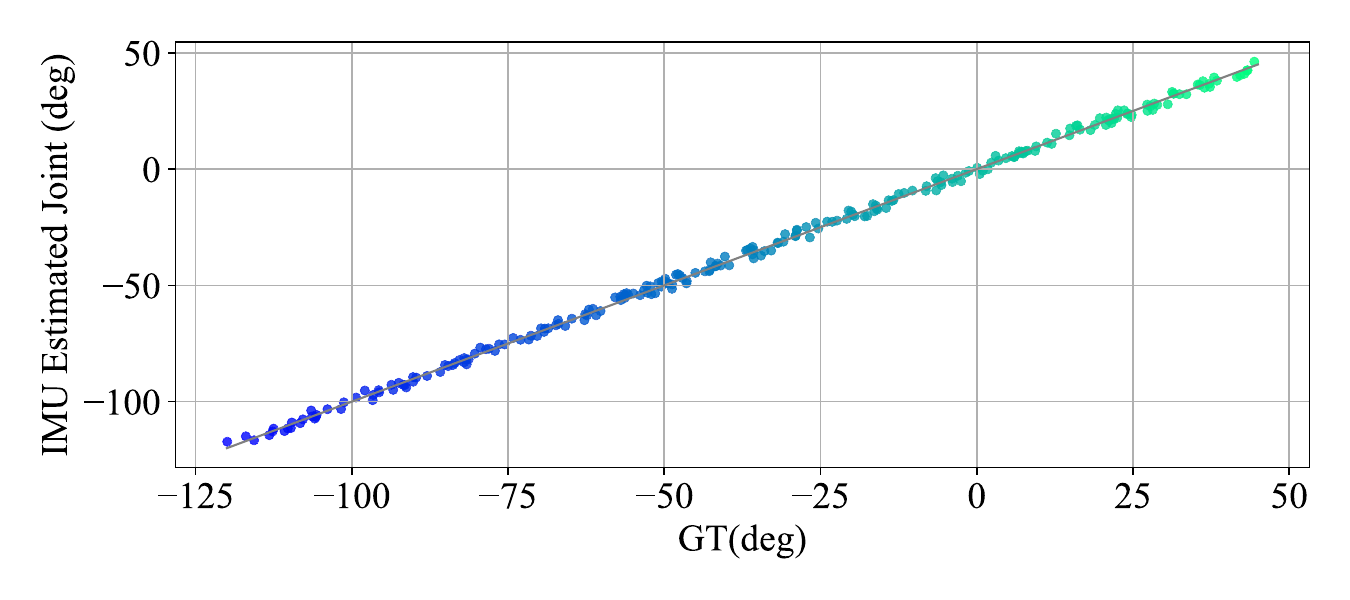}
    \caption{\textbf{Results of single joint measurement.} Our system’s angle output closely matches the measurements from the Nokov MoCap system.}
    \label{fig:experiment_0_result}
\end{figure}

According to our specific use case, it is necessary to validate the accuracy of joint pose alignment using a pair of IMUs. We have designed a 3D-printed base plate to model scenarios of finger movement. As illustrated in Fig.\ref{fig:experiment_0_setting}, the device is divided into two sections, A and B, connected via a rotating axis. Each section has designated positions for mounting motion capture marker balls, facilitating the capture of sections A and B's postures using the Nokov system and enabling the calculation of the true rotation angle of the axis. With Nokov's tracking accuracy at $0.5$ mm, at a scale of $100$ mm, the measurement accuracy for rotation reaches $0.3$ degrees, which meets our requirements. In our experiments, we use the output from the Nokov system as the benchmark to gauge the error in pose estimation provided by the IMU system. As the results shown in Fig.\ref{fig:experiment_0_result}, the bias of the IMU's pose estimation is $\pm 2.7^{\circ}$, with a standard deviation of $\pm 1.8^{\circ}$. The non-linearity of the sensor is 0.7\%. Furthermore, its accuracy does not fluctuate with changes in joint angles. We can deem such an IMU system is reliable.

\begin{figure}
    \centering
    \includegraphics[width=1\columnwidth]{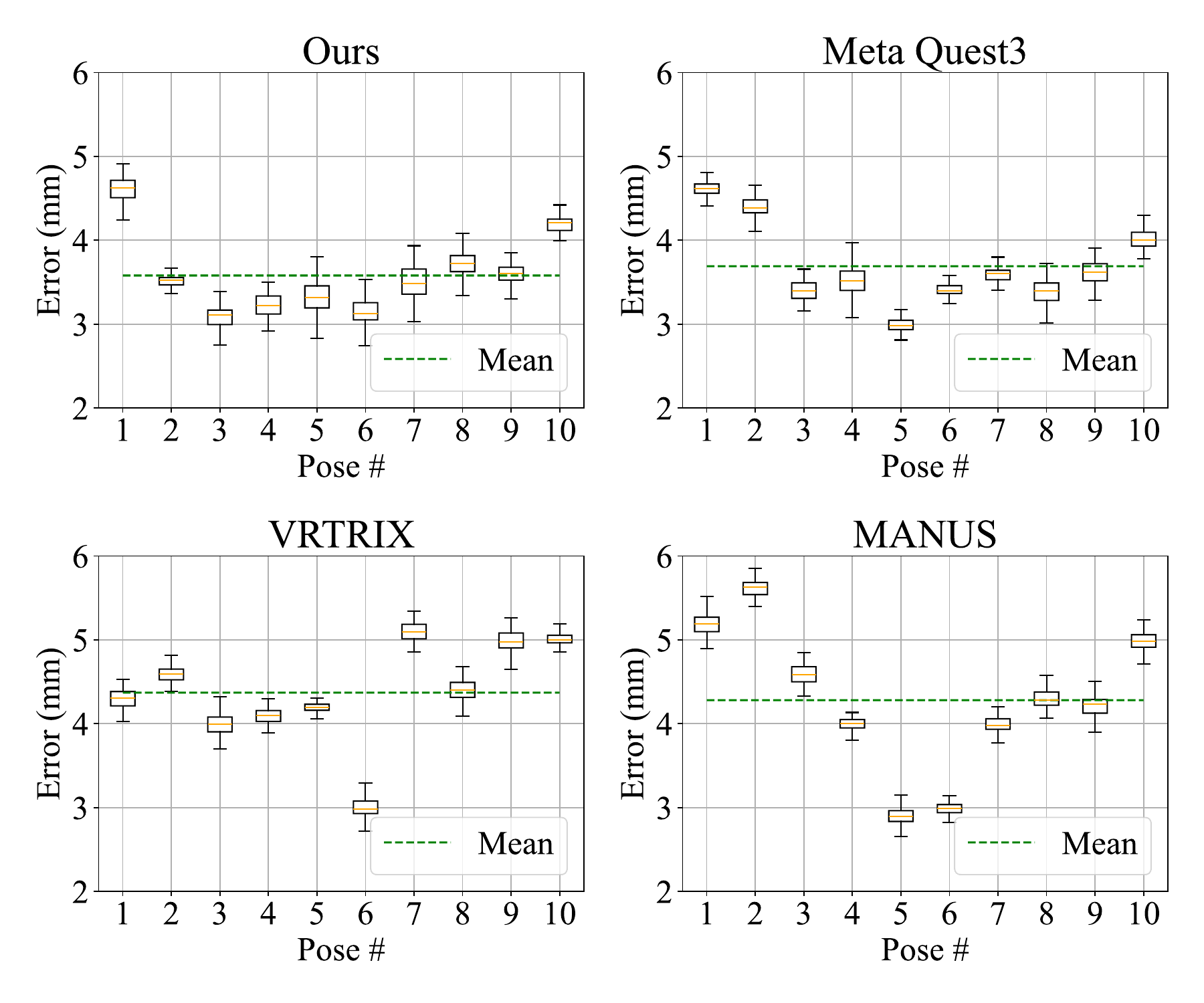}
    \caption{\textbf{Quantitative results of shape reconstruction.} Our solution demonstrates comparable performance to the
    Quest3 and outperforms other commercially available gloves.}
    \label{fig:experiment_1_plot}
\end{figure}

\begin{table}
\centering
\begin{tabular}{l@{\hspace{8pt}}|@{\hspace{8pt}}ccccc}
\toprule
\textbf{Method $\downarrow$} & \multicolumn{5}{c}{\textbf{Distance (mm)}} \\
\cmidrule{2-6}
 & \textbf{Index} & \textbf{Middle} & \textbf{Ring} & \textbf{Little} & \textbf{Average} \\
\midrule
Ours & \textbf{7.2} & 16.5 & \textbf{12.0} & 27.1 & \textbf{15.7} \\
Meta Quest3 & 11.0 & 21.5 & 18.9 & 22.3 & 19.6 \\
VRTRIX & 20.9 & \textbf{14.4} & 25.3 & \textbf{15.1} & 18.9 \\
MANUS & 28.0 & 32.3 & 36.7 & 35.7 & 33.2 \\
\bottomrule
\end{tabular}
\caption{\textbf{Results of fingertip pinch tracking.} Our method achieves the best average error. Notably, it attains the lowest error for the index and ring fingers, confirming its superior accuracy in most finger measurements.}
\label{tab:thumb_distances}
\end{table}

\subsection{Shape Reconstruction}

As a data glove designed for hand motion reconstruction, we are particularly interested in its fundamental shape reconstruction capabilities. To conduct this experiment, a Photoneo MotionCam depth camera was mounted on a tripod and oriented toward a flat white wall, with the space between the camera and the wall cleared to ensure unobstructed imaging. A volunteer wearing the glove performed a series of complex and diverse hand poses. During this process, the Photoneo camera captured a high-precision grayscale depth image of the volunteer’s hand while the glove recorded its reconstruction results concurrently.

\begin{figure*}
    \centering
    \includegraphics[width=0.8\textwidth]{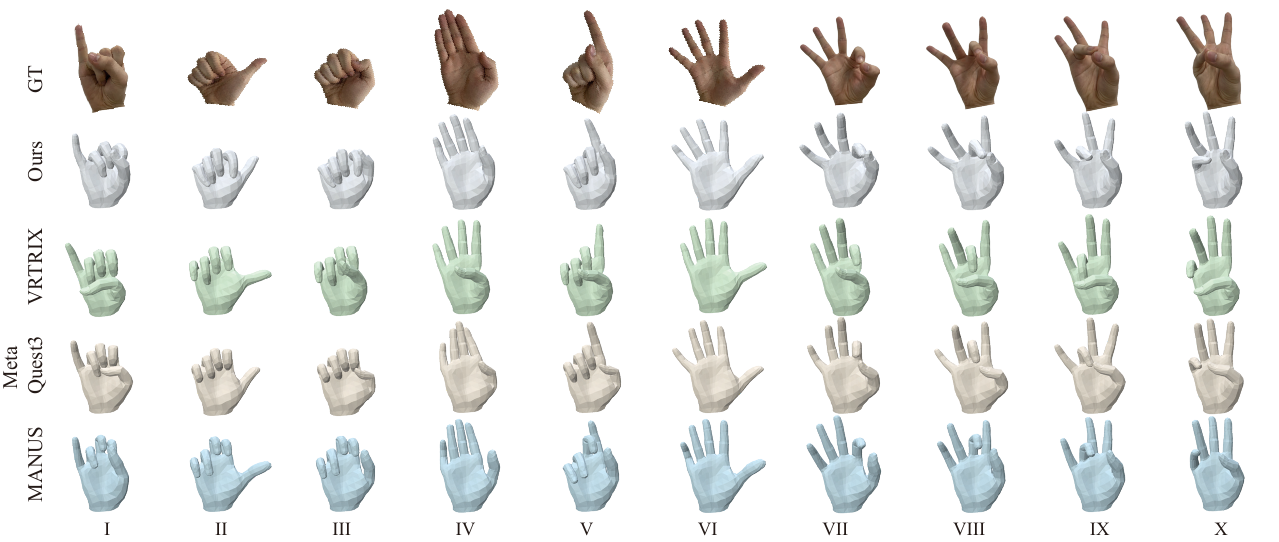}
    \caption{\textbf{Qualitative results of shape reconstruction.}}
    \label{fig:experiment_1_matrix}
\end{figure*}

\begin{table*}
\centering
\begin{tabular}{l@{\hspace{8pt}}|@{\hspace{8pt}}cc@{\hspace{8pt}}cc@{\hspace{8pt}}cc@{\hspace{8pt}}cc@{\hspace{8pt}}cc@{\hspace{8pt}}|c}
\toprule
& \multicolumn{10}{c}{\textbf{Test Result (mm)}} & \textbf{Total Avg.} \\
\cmidrule{2-11}
\textbf{Method $\downarrow$} & \multicolumn{2}{c}{\textbf{\#1}} & \multicolumn{2}{c}{\textbf{\#2}} & \multicolumn{2}{c}{\textbf{\#3}} & \multicolumn{2}{c}{\textbf{\#4}} & \multicolumn{2}{c}{\textbf{\#5}} \\
& Avg. & Std. & Avg. & Std. & Avg. & Std. & Avg. & Std. & Avg. & Std. \\
\midrule
Ours & \textbf{22.6} & \textbf{5.5} & 16.5 & 4.2 & \textbf{17.9} & \textbf{2.9} & 26.6 & \textbf{2.0} & \textbf{17.5} & \textbf{3.3} & \textbf{20.2} \\
Meta Quest3 & 69.7 & 13.0 & 17.7 & \textbf{1.3} & 18.0 & 4.8 & 16.7 & 11.1 & 18.1 & 9.2 & 28.0 \\
VRTRIX & 41.9 & 6.4 & \textbf{14.4} & 2.3 & 19.1 & 3.6 & \textbf{16.2} & 2.3 & 30.8 & 8.2 & 24.5 \\
MANUS & 28.0 & 7.3 & 19.8 & 3.9 & 26.7 & 4.6 & 25.7 & 4.0 & 33.2 & 12.3 & 26.7 \\
\bottomrule
\end{tabular}
\caption{\textbf{Quantitative results of Hand-Object Interaction Consistency Experiment }As can be observed, our proposed method outperforms other approaches on most objects. The relatively lower scores of MANUS and VRTRIX are largely attributable to unnatural hand motions induced by external forces during the interaction, which are consequently filtered by their built-in algorithms.}
\label{tab:thumb_distances2}
\end{table*}

To compare the reconstruction results with the high-precision depth images, the GroundedSAM model \cite{GSAM} was employed to apply a ''hand'' query to the depth image, enabling precise extraction of the volunteer’s hand. The extracted regions were then converted into a point cloud and compared with the glove’s reconstruction results. To account for the Photoneo camera’s output of a partial point cloud, we use the unidirectional Chamfer Distance from the partial point cloud $P$ to the reconstructed hand mesh vertices $V$ as the error metric:
$$
E_{sr} = \frac{1}{\vert P \vert} \sum_{p \in P}\min_{v \in V} \Vert p - v\Vert^2_2
$$
The experiment was repeated 10 times for each pose, and the results were averaged. As summarized in Fig. \ref{fig:experiment_1_plot}, our solution demonstrates comparable performance to the Quest3 and outperforms other commercially available gloves.

However, we find that the Meta Quest3 performs comparably or even better in fingertip pinching. We attribute this to two main factors. First, Quest3 uses visual information to reconstruct a hand mesh, thereby capturing more precise geometric details when subtle contact actions are visible. Second, the finger-mounted dorsal IMUs can collide with each other, especially during challenging poses like thumb-pinky contact, causing positional drift and reducing accuracy.
\subsection{Fingertip Pinch Tracking}

Motion capture gloves are commonly utilized in virtual reality (VR). For example, \cite{unifolding} adopted a commercial motion capture glove in VR to generate a dataset that trains a grasp policy for robot arms. Therefore, minimizing contact error between fingertips is critical, as it directly impacts the accuracy of essential operations such as grasping. To address this, we designed an experiment to quantify fingertip contact error in the \textit{Pinch} pose.  After properly calibrating the glove, participants performed a series of instructed gestures, sequentially bringing the thumb into contact with the other four fingertips. During the experiment, wrist rotation was randomly applied to simulate real-world operating conditions. For each \textit{Pinch} pose, a 10-second sequence of distances between fingertip endpoints was recorded, and the averaged measurements were summarized in Tab. \ref{tab:thumb_distances}. Our solution demonstrated marginally superior performance compared to the VRTRIX glove, whereas the MANUS glove exhibited some deviation. The Quest3’s performance was notably suboptimal due to its helmet-mounted cameras’ inability to capture hand poses when the user’s palm faced downward.

For the \textit{little pinching} pose, VRTRIX outperforms both our glove and other methods by a large margin. This is because VRTRIX systematically tends to bend the thumb across the palm, regardless of the actual thumb position, as shown in Fig.\ref{fig:experiment_1_matrix}, thereby improving its performance in the evaluated metric. However, this behavior can cause the reconstructed mesh to deviate significantly from the ground truth.

\subsection{Hand-Object Interaction Consistency}

In many situations, the human hand is used to manipulate various objects. Therefore, we are particularly interested in whether the recorded hand mesh sequences can faithfully reflect the actual physical processes that occur when gloves interact with objects. To explore this, we selected five objects: $\mathtt{cube\_large}$, $\mathtt{cup}$, $\mathtt{utah\_teapot}$, $\mathtt{stanford\_rabbit}$, and $\mathtt{elephant}$ from the ContactPose \cite{contactpose} dataset, which provides a collection of digital models of common objects, each equipped with mounting points for marker balls.

Following the methods described in \cite{contactpose}, we installed markers on these models to track their position and orientation changes using Nokov. We then adjusted the visualization and recording programs based on the Nokov readings. Our glove setup already includes a tracking solution for the palm, so no modifications were necessary. For VRTRIX, MANUS gloves, and Quest3, we created multiple replicas of the dorsal tracker for palm tracking. For each combination of object and glove, we had volunteers perform operations for approximately 60 seconds, recording the palm and object's position and orientation via Nokov, and hand reconstruction results via custom methods.

Our focus was on the regions of interest in the hand mesh (the tips of each finger) and the interpenetration of the hand mesh with the object mesh. We selected 30 random moments from the sequence, ensuring that all fingers were in contact with the object. For these moments, we calculated the point-to-mesh distance, which is distance from the fingertip endpoint to the nearest point on the object mesh. A lower value indicates a better reconstruction. Since all palm positions and orientations were provided by Nokov and the tracking devices, this allowed for a fair comparison. The experimental results in Tab. \ref{tab:thumb_distances2} indicate that our method performs slightly better than Quest3 and VRTRIX while significantly outperforming MANUS gloves.

\subsection{Drift Analysis}

Two primary sources of drift contribute to the inaccuracy of FSGlove: IMU drift inherent to inertial sensors, and the stretching of the glove fabric.

\paragraph{IMU Drift}
According to the Hi229 IMU datasheet \cite{hi229}, the typical angular error after 30 minutes of operation ranges from 3$^\circ$-10$^\circ$, depending on the fusion mode. However, during the above experiments, which often exceeded 10 minutes, no obvious drift was encountered. In the single joint measurement experiment (see Section \ref{sec:single-joint-measurement}), we observed a standard deviation of $\pm 1.8^{\circ}$. In the hand-object interaction consistency experiment, the standard deviation of FSGlove is on par with other methods (see Table \ref{tab:thumb_distances2}). Therefore, we conclude that IMU drift has an insignificant impact on short-period data collection tasks.

\paragraph{Stretching Effect}
Prolonged use can cause the glove to stretch, invalidating previous calibrations. However, the pose calibration module, as previously discussed in Section \ref{sec:pose_calibration}, makes no assumptions about the initial position or orientation of the IMU modules relative to the phalanxes. For this reason, users are instructed to recalibrate the glove daily to counteract the stretching effect.

\section{Conclusion and Future Work}
This work introduces FSGlove, an open-source, high-DoF (48-DoF) data glove combining inertial sensing with personalized shape calibration, addressing critical gaps in hand motion capture. By integrating densely instrumented IMUs and DiffHCal, a differentiable calibration framework, FSGlove achieves anatomically consistent hand modeling, resolving joint kinematics, shape parameters, and sensor misalignment in a unified workflow. Experiments prove its state-of-the-art joint angle accuracy, robust shape reconstruction, and superior contact fidelity compared to commercial systems. The system’s open design and compatibility with VR and robotics systems democratize access to high-fidelity hand tracking, enabling more possible downstream applications.

Future work includes enhancing IMU precision for sub-degree tracking, extending DiffHCal to adapt to soft-tissue deformation during contact dynamically, and generalizing shape estimation to diverse hand anthropometries. Integrating tactile feedback could further improve interaction realism, while federated learning frameworks might enable personalized calibration without manual intervention. Exploring applications in surgical robotics or immersive teleoperation will validate FSGlove’s scalability. By open-sourcing the design, we invite community-driven advancements, bridging the fidelity gap between human and robotic manipulation.

\addtolength{\textheight}{-12cm}

\bibliographystyle{IEEEtran}
\bibliography{reference}

\begin{thebibliography}{10}
\providecommand{\url}[1]{#1}
\csname url@rmstyle\endcsname
\providecommand{\newblock}{\relax}
\providecommand{\bibinfo}[2]{#2}
\providecommand\BIBentrySTDinterwordspacing{\spaceskip=0pt\relax}
\providecommand\BIBentryALTinterwordstretchfactor{4}
\providecommand\BIBentryALTinterwordspacing{\spaceskip=\fontdimen2\font plus
\BIBentryALTinterwordstretchfactor\fontdimen3\font minus
  \fontdimen4\font\relax}
\providecommand\BIBforeignlanguage[2]{{%
\expandafter\ifx\csname l@#1\endcsname\relax
\typeout{** WARNING: IEEEtran.bst: No hyphenation pattern has been}%
\typeout{** loaded for the language `#1'. Using the pattern for}%
\typeout{** the default language instead.}%
\else
\language=\csname l@#1\endcsname
\fi
#2}}

\bibitem{cyberglove}
CyberGlove, ``Cyberglove iii introduction,''
  \url{https://www.cyberglovesystems.com/cyberglove-iii/}, 2025, accessed:
  March 1, 2025.

\bibitem{leaphand}
K.~Shaw, A.~Agarwal, and D.~Pathak, ``Leap hand: Low-cost, efficient, and
  anthropomorphic hand for robot learning,'' \emph{Robotics: Science and
  Systems (RSS)}, 2023.

\bibitem{manus}
Manus, ``Manus glove product introduction,''
  \url{https://www.manusmachina.com}, 2025, accessed: March 1, 2025.

\bibitem{vrtrix}
VRTrix, ``Vrtrix glove product introduction,'' \url{https://www.vrtrix.com},
  2025, accessed: March 1, 2025.

\bibitem{glove1}
S.~Zhu, A.~Stuttaford-Fowler, A.~Fahmy, C.~Li, and J.~Sienz, ``Development of a
  low-cost data glove using flex sensors for the robot hand teleoperation,'' in
  \emph{2021 3rd International Symposium on Robotics \& Intelligent
  Manufacturing Technology (ISRIMT)}, 2021, pp. 47--51.

\bibitem{glove2}
B.-S. Lin, I.-J. Lee, P.-Y. Chiang, S.-Y. Huang, and C.-W. Peng, ``A modular
  data glove system for finger and hand motion capture based on inertial
  sensors,'' \emph{Journal of Medical and Biological Engineering}, vol.~39, pp.
  532--540, 2019.

\bibitem{glove3}
B.~O'Flynn, J.~Torres, J.~Connolly, J.~Condell, K.~Curran, and P.~Gardiner,
  ``Novel smart sensor glove for arthritis rehabiliation,'' in \emph{2013 IEEE
  International Conference on Body Sensor Networks}, 2013, pp. 1--6.

\bibitem{mano}
J.~Romero, D.~Tzionas, and M.~J. Black, ``Embodied hands: Modeling and
  capturing hands and bodies together,'' \emph{ACM Transactions on Graphics,
  (Proc. SIGGRAPH Asia)}, vol.~36, no.~6, Nov. 2017.

\bibitem{sayreglove}
D.~Sturman and D.~Zeltzer, ``A survey of glove-based input,'' \emph{IEEE
  Computer Graphics and Applications}, vol.~14, no.~1, pp. 30--39, 1994.

\bibitem{dataglove}
H.~Eglowstein, ``Reach out and touch your data,'' \emph{Byte}, vol.~15, no.~7,
  pp. 283--290, July 1990.

\bibitem{powerglove}
D.~L. Gardner, ``The power glove,'' \emph{Des. News}, vol.~45, pp. 63--68, Dec.
  1989.

\bibitem{glove_survey}
L.~Dipietro, A.~M. Sabatini, and P.~Dario, ``A survey of glove-based systems
  and their applications,'' \emph{IEEE Transactions on Systems, Man, and
  Cybernetics, Part C (Applications and Reviews)}, vol.~38, no.~4, pp.
  461--482, 2008.

\bibitem{resistiveglove}
J.~Hughes, A.~Spielberg, M.~Chounlakone, G.~Chang, W.~Matusik, and D.~Rus, ``A
  simple, inexpensive, wearable glove with hybrid resistive-pressure sensors
  for computational sensing, proprioception, and task identification,''
  \emph{Advanced Intelligent Systems}, vol.~2, no.~6, p. 2000002.

\bibitem{flexglove}
S.~Zhu, A.~Stuttaford-Fowler, A.~Fahmy, C.~Li, and J.~Sienz, ``Development of a
  low-cost data glove using flex sensors for the robot hand teleoperation,'' in
  \emph{2021 3rd International Symposium on Robotics \& Intelligent
  Manufacturing Technology (ISRIMT)}, 2021, pp. 47--51.

\bibitem{imuglove1}
B.~Fang, F.~Sun, H.~Liu, and D.~Guo, ``A novel data glove for fingers motion
  capture using inertial and magnetic measurement units,'' in \emph{2016 IEEE
  International Conference on Robotics and Biomimetics (ROBIO)}, 2016, pp.
  2099--2104.

\bibitem{imuglove2}
Q.~Liu, G.~Qian, W.~Meng, Q.~Ai, C.~Yin, and Z.~Fang, ``A new immu-based data
  glove for hand motion capture with optimized sensor layout,''
  \emph{International Journal of Intelligent Robotics and Applications},
  vol.~3, no.~1, pp. 19--32, March 2019, (c) 2019, Springer Nature Singapore
  Pte Ltd. This is an author produced version of a paper published in the
  International Journal of Intelligent Robotics and Applications. Uploaded in
  accordance with the publisher's self-archiving policy.

\bibitem{lin_design_2018}
B.-S. Lin, I.-J. Lee, S.-Y. Yang, Y.-C. Lo, J.~Lee, and J.-L. Chen,
  ``\BIBforeignlanguage{en}{Design of an {Inertial}-{Sensor}-{Based} {Data}
  {Glove} for {Hand} {Function} {Evaluation}},''
  \emph{\BIBforeignlanguage{en}{Sensors}}, vol.~18, no.~5, p. 1545, May 2018.

\bibitem{calibration1}
T.-S. Chou and A.~Gadd, ``Hand-eye: A vision-based approach to data glove
  calibration,'' 2000.

\bibitem{calibration2}
F.~Kahlesz, G.~Zachmann, and R.~Klein, ``'visual-fidelity' dataglove
  calibration,'' in \emph{Proceedings Computer Graphics International, 2004.},
  2004, pp. 403--410.

\bibitem{calibration3}
Z.~Sun, G.~Bao, J.~Li, and Z.~Wang, ``Research of dataglove calibration method
  based on genetic algorithms,'' in \emph{2006 6th World Congress on
  Intelligent Control and Automation}, vol.~2, 2006, pp. 9429--9433.

\bibitem{calibration4}
J.~Zhou, F.~Malric, and S.~Shirmohammadi, ``A new hand-measurement method to
  simplify calibration in cyberglove-based virtual rehabilitation,'' \emph{IEEE
  Transactions on Instrumentation and Measurement}, vol.~59, no.~10, pp.
  2496--2504, 2010.

\bibitem{calibration5}
J.~Connolly, J.~Condell, K.~Curran, and P.~Gardiner, ``Improving data glove
  accuracy and usability using a neural network when measuring finger joint
  range of motion,'' \emph{Sensors}, vol.~22, no.~6, 2022.

\bibitem{hi229}
S.~L. Tech, ``Hi229 9-axis imu/vru/ahrs,''
  \url{https://sealandtech.com.tw/en/product_anrot_hi229.html}.

\bibitem{rfu}
H.~Fu*, W.~Xu*, R.~Ye*, H.~Xue, Z.~Yu, T.~Tang, Y.~Li, W.~Du, J.~Zhang, and
  C.~Lu, ``Demonstrating rfuniverse: A multiphysics simulation platform for
  embodied ai,'' in \emph{{RSS} Robotics: Science and Systems}, 2023.

\bibitem{harp}
K.~Karunratanakul, S.~Prokudin, O.~Hilliges, and S.~Tang, ``Harp: Personalized
  hand reconstruction from a monocular rgb video,'' in \emph{Proceedings of the
  IEEE/CVF conference on computer vision and pattern recognition}, 2023, pp.
  12\,802--12\,813.

\bibitem{GSAM}
T.~Ren, S.~Liu, A.~Zeng, J.~Lin, K.~C. Li, H.~Cao, J.~Chen, X.~Huang, Y.~Chen,
  F.~Yan, Z.~Zeng, H.~Zhang, F.~Li, J.~Yang, H.~Li, Q.~Jiang, and L.~Zhang,
  ``Grounded sam: Assembling open-world models for diverse visual tasks,''
  \emph{arXiv preprint arXiv:2401.14159}, 2024.

\bibitem{unifolding}
H.~Xue, Y.~Li, W.~Xu, H.~Li, D.~Zheng, and C.~Lu, ``Unifolding: Towards
  sample-efficient, scalable, and generalizable robotic garment folding,'' in
  \emph{Conference on Robot Learning}.\hskip 1em plus 0.5em minus 0.4em\relax
  PMLR, 2023, pp. 3321--3341.

\bibitem{contactpose}
S.~Brahmbhatt, C.~Tang, C.~D. Twigg, C.~C. Kemp, and J.~Hays, ``Contactpose: A
  dataset of grasps with object contact and hand pose,'' in \emph{16th European
  Conference on Computer Vision (ECCV 2020)}, 2020, pp. 361--378.

\end{thebibliography}

\end{document}